\documentclass[9pt,twocolumn,twoside]{pnas-new}

\templatetype{pnasresearcharticle} %
\usepackage{times}
\usepackage{latexsym}

\usepackage[T1]{fontenc}

\usepackage[utf8]{inputenc}

\usepackage{microtype}

\usepackage{inconsolata}

\usepackage{graphicx}

\usepackage{amsmath}

\usepackage{url}
\usepackage{amsthm}
\usepackage{amsfonts}
\usepackage{bm}
\usepackage{thmtools,mathtools}
\usepackage{thm-restate}
\usepackage{booktabs}
\usepackage{multirow}
\usepackage{booktabs}
\usepackage{verbatim}
\usepackage{mdwlist}
\usepackage{dsfont}
\usepackage{subcaption}
\usepackage{array}
\usepackage{pifont}
\usepackage{xspace}

\usepackage{multicol}
\usepackage{float}
\usepackage{lipsum}
\usepackage[font=small]{caption}
\usepackage{svg}
\usepackage{wrapfig}

\theoremstyle{definition}

\makeatletter
\newcommand{\removelatexerror}{\let\@latex@error\@gobble}
\makeatother

\newcommand{\codename}{\textit{Sui Generis}\xspace}

\begin{document}

\title{Echoes in AI: Quantifying Lack of Plot Diversity in LLM Outputs}

\author[a,1]{Weijia Xu}
\author[a]{Nebojsa Jojic}
\author[a]{Sudha Rao}
\author[a]{Chris Brockett}
\author[a]{Bill Dolan}

\affil[a]{Microsoft Research, Redmond, WA 98052, USA}

\leadauthor{Xu}

\significancestatement{Reading through a set of texts generated by LLMs under the same prompt can be a disappointing experience. The first might seem convincingly like a novel, creative work by a human writer. Read more, though, and the lack of diversity in these LLM-generated outputs reveals itself. We show that short stories generated in this way often contain repetitive combinations of plot elements, while human-written stories maintain a higher level of uniqueness. This research is the first to quantify this observation, and to introduce an automatic metric aimed at measuring the usefulness of LLMs for creative content generation at the narrative level. We believe that this metric will help drive progress towards LLMs that generate more diverse and creative content.}

\authorcontributions{Author contributions: W.X., N.J., S.R., C.B., and B.D. designed research; W.X. and S.R. performed research; W.X., N.J., S.R., C.B., and B.D. analyzed data and results; and W.X., N.J., S.R., C.B., and B.D. wrote the paper.}
\authordeclaration{The authors declare no competing interest.}
\correspondingauthor{\textsuperscript{1}To whom correspondence should be addressed. E-mail: weijiaxu@microsoft.com}

\keywords{LLM $|$ text generation $|$ deep learning $|$ creativity $|$ AI}

\begin{abstract}
With rapid advances in large language models~(LLMs), there has been an increasing application of LLMs in creative content ideation and generation. A critical question emerges: can current LLMs provide ideas that are diverse enough to truly bolster collective creativity? We examine two state-of-the-art LLMs, GPT-4 and LLaMA-3, on story generation and discover that LLM-generated stories often consist of plot elements that are echoed across a number of generations. To quantify this phenomenon, we introduce the \codename score, the first automatic metric that measures the uniqueness of a plot element among alternative storylines generated using the same prompt under an LLM. Evaluating on 100 short stories, we find that LLM-generated stories often contain combinations of idiosyncratic plot elements echoed frequently across generations and across different LLMs, while plots from the original human-written stories are rarely recreated or even echoed in pieces. 
Moreover, our human evaluation shows that the ranking of \codename scores among story segments correlates moderately with human judgment of surprise level, even though score computation is completely automatic without relying on human judgment.
\end{abstract}

\dates{Edited by Jeffrey Ullman, Stanford University, Stanford, CA; received March 17, 2025; accepted July 9, 2025}
\doi{\url{www.pnas.org/cgi/doi/10.1073/pnas.2504966122}}

\maketitle
\thispagestyle{firststyle}
\ifthenelse{\boolean{shortarticle}}{\ifthenelse{\boolean{singlecolumn}}{\abscontentformatted}{\abscontent}}{}

\firstpage[1]{5}

\section{Introduction}

\looseness=-1
Rapid advances in large language models~(LLMs) have spurred an ongoing debate on the usefulness of these models for tasks that require human-level creativity. On the one hand, there are works that highlight the strengths of LLMs in creative writing \cite{bellemarepepin2024divergentcreativityhumanslarge, orwig2024language}, poetry generation \cite{porter2024ai}, idea generation \cite{lee2024empirical,si2024can} and even creative thinking \cite{GUZIK2023100065}. On the other hand, there has been research arguing that LLM creativity is much weaker than human creativity \cite{Chakrabarty2024} and that LLM-generated stories are identifiably bad \cite{Sato.2023,Levenson.2023}. 

One recent study 
\cite{doi:10.1126/sciadv.adn5290}
finds that while the use of an AI assistant in writing appears to enhance the creativity of individual writers, it also reduces the collective diversity of novel content produced by multiple writers. Similarly, a user study on argumentative essay writing \cite{padmakumar2024does} finds that writing with LLMs reduces the diversity of content produced by a group of users. These findings reflect a phenomenon that is by now familiar\footnote{\url{https://edtechrce.org/educators-develop-strategies-to-detect-ai-generated-student-work/}} 
to teachers whose students use LLMs for help with writing assignments: while any individual LLM output might seem compelling and novel, reading through multiple texts produced by the same prompt can be a deflating experience. A narrative element that seems strikingly innovative and creative when encountered in the first output may begin to seem, by the time it is encountered in~10 more outputs, more like the product of a deterministic process than human-like creativity.

\looseness=-1

An example of this phenomenon is shown in Table \ref{tab:introductory_continuation_examples}, which shows a small subset of 100 short story continuations generated by GPT-4\footnote{We use temperature~$\tau=1$ as LLMs are trained with the goal of fully capturing the data distribution with~$\tau=1$.} given the first part of \textit{Give It Up}, a short story by Franz Kafka\footnote{\url{https://www.flashfictiononline.com/article/give-it-up/}} (see the full list of LLM-generated continuations in supplementary material). None of GPT-4's continuations resemble Kafka's own ending for the story, in which a policeman disconcertingly tells the direction-seeking narrator to ``Give it up!'' and abruptly turns away. Instead, in 50 out of 100 generations, the policeman gives instructions to take the second left; in 18/100 to take the second right; and in others walks with the protagonist to show them the way. In 16/100 of the generations, a bakery is mentioned as a landmark. %
These \textit{ echoes}, as we refer to them here, repeat surprisingly frequently across generations at all semantic levels and not necessarily in the same order. The previously observed reduction in lexical diversity of LLM outputs \cite{mohammadi2024creativityleftchatprice} corresponds to what we view as semantically lower-level echoes, such as the frequent use of word ``bakery.'' However, lexical diversity cannot be used to explain or quantify echoes at the higher semantic level. Narrative developments such as the policeman's decision to show the way or point the protagonist towards a specific direction
can be paraphrased in different ways, as the example continuations show. These more abstract echoes are not merely the result of repeated tokens.

In this paper, we aim to quantify such repetitions at the \emph{narrative} level by introducing a \codename score.\footnote{Sui Generis is a Latin phrase meaning ``of its own kind,'' used in English biology and law literature to indicate something unlikely to be repeated or recreated.}
We use story generation as a testbed. To compute the \codename score for a story, we first ask the LLM to generate many alternative continuations for the same story given varying lengths of the story prefix as context. We then count the number of times an original story segment is echoed~(at the narrative level) in the alternative continuations. We adopt the intuition that a segment is more likely to appear in other story samples and is thus less unique if it is echoed in a larger number of alternative continuations given less context from the previous plot.

\looseness=-1
We test the \codename score on 100 stories consisting of around~$3,700$ segments from two datasets: WritingPrompts, which contains short stories posted on Reddit; and a set of television episode plot summaries from Wikipedia. We experiment with two state-of-the-art LLMs: GPT-4 and LLaMA-3. Our results highlight the lack of diversity in LLM outputs: LLM-generated stories are often composed of the same combinations of idiosyncratic plot elements that are echoed frequently across its own generations and even across different LLMs, while only rarely exhibiting plot elements found in human narratives sparked by the same prompts. As the \codename score is automatically calculated without human judgment, this result provides a quantitative explanation for the previous qualitative findings that the use of an AI assistant appears to reduce output collective diversity~\cite{doi:10.1126/sciadv.adn5290}. It explains the contradictory findings in previous user studies on whether LLM generations have reached human-level creativity, as our study suggests that an LLM generation may seem novel to a human who has not seen many of its generations in that domain, but may seem banal to someone more familiar with its outputs. 
Furthermore, our human evaluation of the surprise level of story segments shows that the \codename score correlates moderately with the average human judgment of surprisal. We observe that the story segments with high scores often correspond to key plot elements or interesting turning points, whereas those with low scores correspond to developments that are bland or highly predictable from the context. 
\begin{table}[!ht]
\begin{tabular}{p{0.93\linewidth}}
\toprule
\small
{\bf Prompt:} It was very early in the morning, the streets clean and deserted, I was walking to the station.
[...]  %
I did not yet know my way very well in this town;
luckily, a policeman was nearby, I ran up to him and breathlessly asked him the way. 
He smiled and said: ``From me you want to know the way?'' ``Yes,'' I said, ``since I cannot find it myself.''\\
\midrule
\small
{\bf Original Continuation:} ``Give it up! Give it up,'' he said, and turned away with a sudden jerk, like people who want to be alone with their laughter.\\
\midrule
\small
{\bf Continuation 1:}  ``Follow this street for two blocks, then take a left at the bakery, and you'll see the station just ahead,'' he instructed with a gentle nod.  \\
\midrule
\small
{\bf Continuation 2:}  ``Keep going straight for two blocks, then take a left and you'll see the station right ahead,'' he explained, giving a friendly nod.\\
\midrule
\small
{\bf Continuation 3:}  He nodded in understanding, his smile broadening as he pointed down the street, ``Take the second left and cross the square; you'll see the station right ahead.''\\
\midrule
\small
{\bf Continuation 4:}  ``Go straight ahead and take your second left, pass the old library, and the station will be right before your eyes,'' he responded, tipping his cap.\\
\midrule
\small
{\bf Continuation 5:} ``Alright,'' he said, his smile broadening, ``I'll guide you. Follow me.'' Grateful, I trailed behind him, my earlier panic subsiding with each step.\\ 
\midrule
\caption{Continuations generated by GPT-4 when prompted with the first part of a short story by Kafka.}
\label{tab:introductory_continuation_examples}
\end{tabular}
\vspace{-30pt}
\end{table}

\section{Related Work}

LLMs are increasingly used in creative writing~\citep{kobak2024delvingchatgptusageacademic,lee2024empirical}. However, there is a debate on whether LLMs boost creativity. Some studies have suggested that LLM-generated content is considered more creative or preferred by users. 
For example, \cite{Kefford.2023} finds that ideas generated by ChatGPT were more likely to be purchased than those generated by Wharton MBA students. \cite{lee2024empirical} find that when participants were asked to generate creative ideas for everyday purposes, the use of ChatGPT increased their creativity.
On the other hand, \cite{begus2023experimentalnarrativescomparisonhuman} finds that AI-generated narratives tend to be less imaginative and only occasionally include plot twists. \cite{Chakrabarty2024} invite expert writers to rate the stories generated by LLMs versus professional writers based on Torrance Test of Creative Writing and discover that LLM-generated stories are less creative than those of professionals.

While these works all focus on evaluating the creativity or novelty of each individual LLM output, \cite{padmakumar2024does} and \cite{doi:10.1126/sciadv.adn5290} discover that, although writers report an AI assistant being helpful in their creative writing, it reduces the collective diversity of content produced by multiple writers. 
Similarly, \cite{Anderson2024Homogenization} shows that ideas generated with the assistance of
LLMs are as diverse as the ideas generated without LLMs at the individual level, but are significantly less diverse at the group level.
These findings suggest that we should examine the \emph{distribution} of LLM creations for a \emph{given} prompt instead of each creation individually.

\begin{figure*}[!ht]
    \centering
    \includegraphics[width=.98\textwidth]{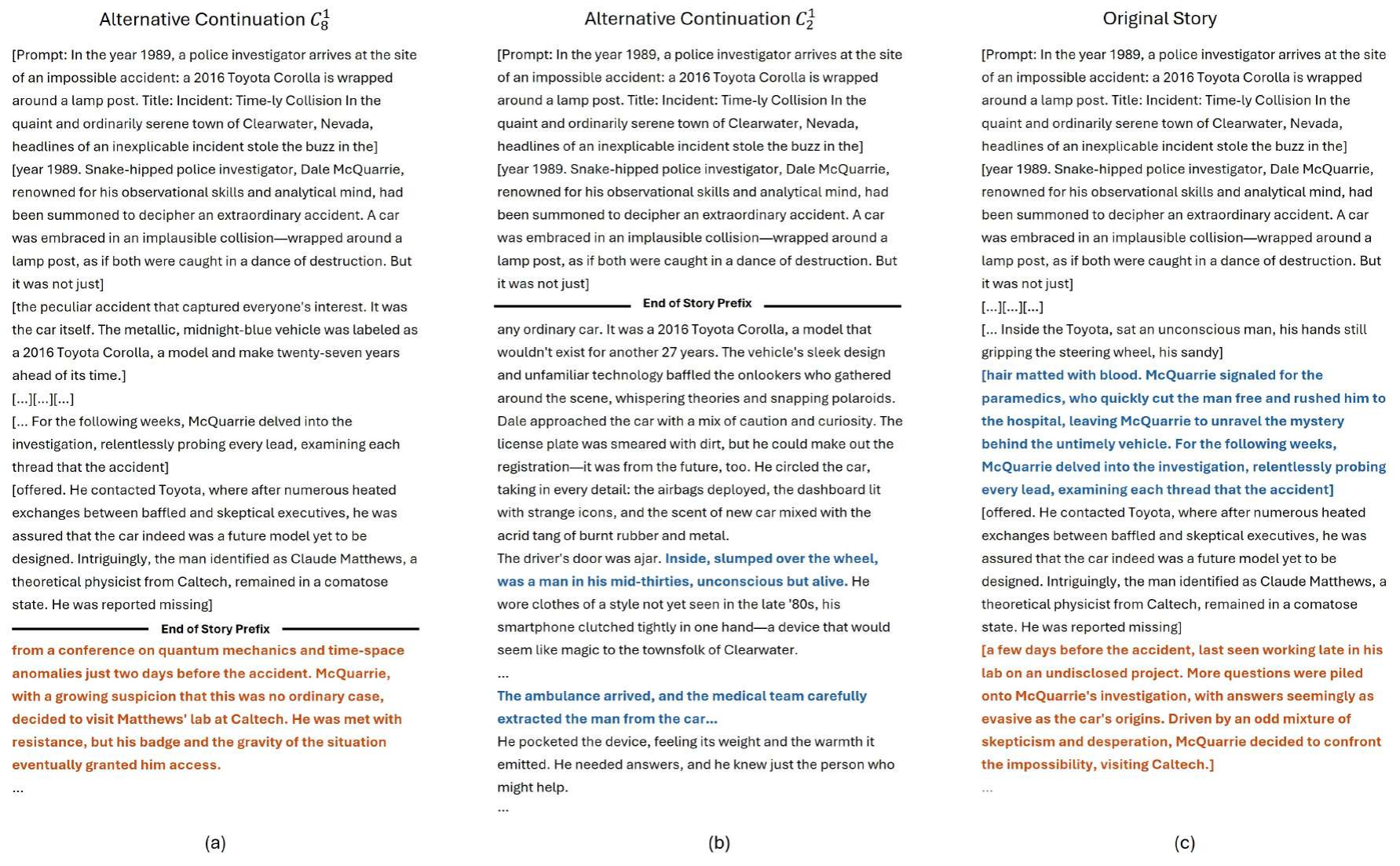}
\caption{Two continuations of the same story with (a) longer prefix, with $j=8$ segments, and (b) shorter prefix, with $j=2$ segments from the same original story, partially shown in (c). Segments are delineated with []. We highlight two segments in (c), $i=7$ in blue and $i=9$ in red, that are echoed in these alternative continuations. The red one~(visiting a scientist’s lab) is echoed only in continuations conditioned on a long story prefix, while the blue one~(discovering a driver inside the car) is echoed frequently even given a short prefix. The \codename score more severely penalizes the echoes discovered given shorter prefix (indicating that similar plot is more likely to be repeated by the LLM).}
\label{fig:echo_example}
\end{figure*}

\looseness=-1
A recent work on detecting novelty in LLM outputs suggests that ``the text must not have been copied from the training data'' \cite{mccoy-etal-2023-much}. This definition appears too narrow and surface-level: nothing resembling the original Kafka text is found in the story continuations by GPT-4~(Table~\ref{tab:introductory_continuation_examples}). Yet by this standard, the generated continuations in story must be treated as novel, even though to human readers it is the original Kafka's ending that is unusual while the generated continuations are quite conventional and lack diversity. 
Other existing metrics that measure diversity in LLM outputs focus either on lexical features, such as ngram overlap between output texts~\cite{shaib2024standardizingmeasurementtextdiversity} or key-point summarizations of the texts~\cite{padmakumar2024does}, or topic-level features, such as the semantic distance between document embeddings~\cite{doi:10.1126/sciadv.adn5290}.
However, lexical features are insufficient for measuring diversity, as \cite{ghosal-etal-2022-novelty} observe, ``identifying novel text is not straightforward because the text may have less lexical overlap yet convey the same information,'' while topic-level features are too high-level to assess the novelty of creative works in which novelty may reside in detailed plots. Thus, in our work, we look beyond lexical or topic-level diversity and introduce the \codename score to measure the uniqueness of text spans at the narrative level.

\section{Method}
\label{sec:metric}

We evaluate the uniqueness of a story segment based on the ``alternative continuations'' generated by LLMs themselves.
Formally, in a story (string) $S$ segmented into $n$ segments $S= (s_1, s_2, ..., s_n)$~(as shown in Figure~\ref{fig:echo_example}(c) with square brackets), at any point~$j \in [1, n-1]$, we truncate the story to its prefix $S_{-j}=(s_1, s_2,...,s_j)$ and consider replacing suffix $S_j=(s_{j+1}, s_{j+2},...)$ by sampling  possible alternative continuations $C_j$ from the model $K$ times:
\begin{equation}
    C_j^k \sim p_{LLM}(\cdot \,|\, S_{-j}), \quad k\in\{1,...,K\}.
    \label{eq:continuations}
\end{equation}
Each string $(S_{-j},C_j^k)$ is thus a full story with its first part taken from the given story S (either human-written or LLM-generated) and the second part being one possible way an LLM would finish it. Figure~\ref{fig:echo_example} shows examples of such continuations with different prefixes: with~(a)~$j=8$ and~(b)~$j=2$ segments from the analyzed story in (c).

\begin{figure}[t]
    \centering
    \includegraphics[width=.48\textwidth]{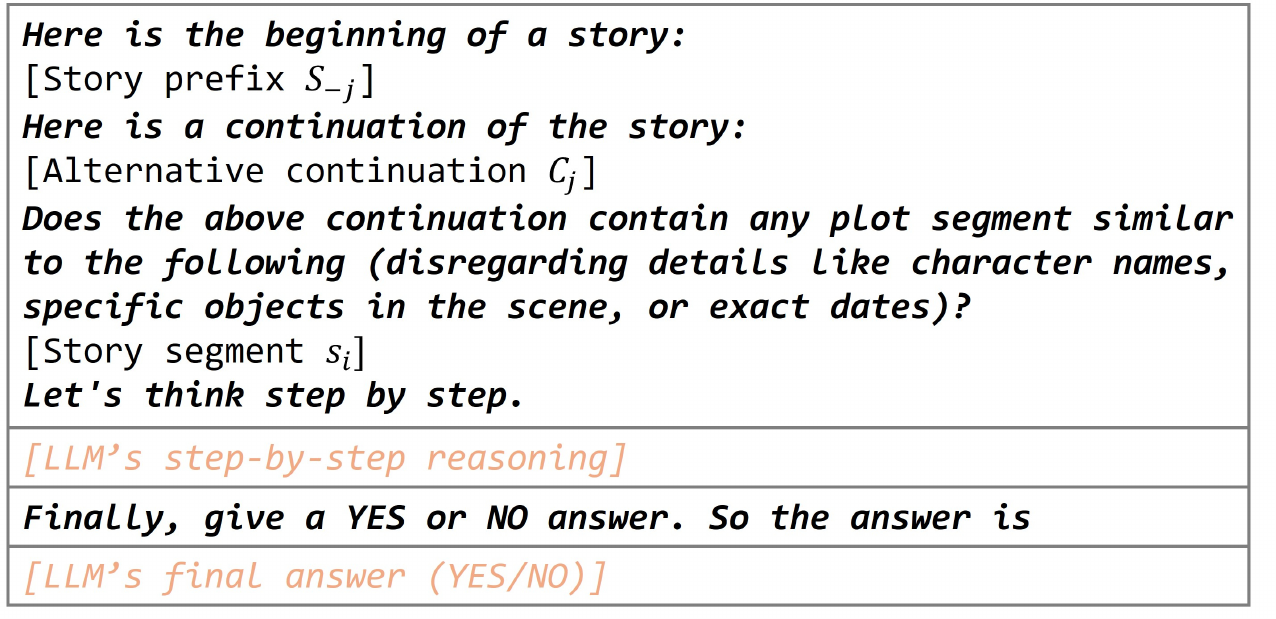}
\caption{Prompt template used for estimating function~$a(s_i, C_j)$, i.e. if the plot in segment~$s_i$ is present in continuation~$C_j$. The texts in black are part of the prompt while the texts in pink should be generated by the LLM.}
\label{fig:judgment_prompt}
\end{figure}

Next, to evaluate how often segment~$s_i$ is repeated across generations, we compute its \textit{echo} scores by comparing it to the alternative continuations~$\{C_1^k\}, \{C_2^k\}, ..., \{C_{i-1}^k\}$ generated from varying lengths of story prefix~$S_{-1}, S_{-2}, ..., S_{-(i-1)}$. Specifically, we compute the \textit{echo} score~$p_{i, j}$, which is an estimated likelihood that similar plot in segment~$s_i$ appears in an alternative continuation:
\begin{equation}
    p_{i, j} \approx \frac{1}{K}\sum_k  a(s_i, C_j^k)
\end{equation}
where~$a(s,C)$ is a binary function that indicates if segment $s$ or its analog is present in continuation $C$. 
\begin{table*}[t!]
\centering
\begin{tabular}{r|m{0.9\textwidth}}
Score & Story Segment \\
\midrule
& {\bf Prompt:} In the middle of the night, a piece of paper is slipped under your door. It says, ``DUCK!'' \\
-- & Confused, you grab the paper and open the door . Looking down the apartment complex corridor you see more and more doors open with perplexed heads poking out one by \\
13.82 & one . No trace is left of whoever put these papers there, they just seem to have appeared all at once . ``Duck?'' Asks that one neighbour that you have rarely ever spoken to . ``Yeah I got it too,'' you reply, ``What do you think it means?'' ``Dunno'' he \\
3.00 & replies as you both turn towards the now gathering crowd . You head towards the crowd, which is incredibly loud with chatter now . ``What's the big deal?'' ``Why should we even care?'' Everyone's thinking the same thing yet everyone seems inexplicably interested . Amidst the crowd you notice that \\
13.82 & the last door is still closed . You point in the direction and immediately the crowd goes silent . Everyone begins to walk over to the closed door when the knob turns . It's dark inside . All you can see is a dark figure inside.As it creeps forward you \\
13.82 & notice an eerie grin . A woman creeps out of the dark room . She looks up, straight into your eyes . Your heart beats faster and faster as she holds up her paper . It reads: ``Goose.'' \\\midrule
\end{tabular}
\caption{An example of a human-written story with both high-scored and low-scored segments. Scores were computed using GPT-4. The story contains a twist ending which is scored high. The high-scored segments correspond to the key plot elements and turning points, while the low-scored one is just an extension of the previous plot.}
\label{tab:mixed_score_human_examples}
\end{table*}
For instance, in Figure~\ref{fig:echo_example}, both $a(s_7, C_8^1)$ and $a(s_9, C_2^1)$ should be~1.
We automate this function by prompting GPT-4 using the prompt template shown in Figure~\ref{fig:judgment_prompt}, as our human evaluation shows that GPT-4's judgment correlates well with human judgment on this task~(see Section~\ref{sec:llm_setup}).\footnote{Note that the granularity of the function~$a(s,C)$ can be easily changed to capture lower-level repetitions~(e.g. the policeman often suggests to take the second left in various continuations in Table~\ref{tab:introductory_continuation_examples}) by adapting the prompt template.}

The \textit{echo} scores~$p_{i, j}$~thus signal how likely and how early in the story~(indicated by position~$j$) the plot in segment~$s_i$ is suggested to be generated. We note that the earlier~$j$ is, the less likely it is that such plot element will be repeated by the LLM in an alterative continuation, therefore echoes induced earlier should be weighted more heavily. Thus, we compute the \codename score of segment~$s_i$ by taking the weighted average of the negative log of echoes~$\{ -\log p_{i, 1}, -\log p_{i, 2}, ..., -\log p_{i, i-1} \}$ where we give higher weights to~$-\log p_{i, j}$ with smaller~$j$'s:
\begin{equation}
    \text{SG}_i = \frac{-\sum_{j=1}^{i-1} \lambda^j \cdot \log p_{i, j}}{\sum_{j=1}^{i-1} \lambda^j}
\label{eq:echo_score}
\end{equation}
where~$\lambda < 1$ is a constant that controls the exponential weight decay. Intuitively, the echoes that are created given shorter prefixes~(e.g. the echo highlighted in Figure~\ref{fig:echo_example}(b)) influence the final score more than the echoes produced given longer prefixes~(e.g. the echo highlighted in Figure~\ref{fig:echo_example}(a)).

The described procedure for computing \codename scores involves neither human judgment nor a large database for matching the generated text. All that is needed are the LLM's own generations and its own judgment of whether a story segment or its equivalent is contained in an alternative story. Table~\ref{tab:mixed_score_human_examples} shows a story with its segment-level \codename scores.
We show in the experiments that, compared to human-written stories, segments of LLM-generated stories have lower scores; in other words, these plot segments tend to be generated repetitively by the LLM, though possibly in a different order or in different parts of the story.

A qualitative user study involving expert writers \cite{Chakrabarty2024} indicated that LLM-generated stories may differ from human-written ones in narrative pacing: In human-written stories, a plot twist is usually carefully foreshadowed and developed in order to maintain tension, which is in line with the tendency for uniform information density in language production~\cite{jaeger2010redundancy}. By contrast, we observe that LLM-generated stories often rapidly accelerate over time without fully resolving the plot.

\looseness=-1
Armed with the \codename score, we can verify this observation quantitatively. Specifically, given the story segments~$s_{1...n}$ through time~$1$ to~$n$, we take the \codename scores~$\text{SG}_{1...n}$ and compute the ``drop ratio'' between the scores of consecutive segments clipped by a threshold~$\theta$:
\begin{equation}
    \text{drop}_i = \max(\frac{\text{SG}_i - \text{SG}_{i+1}}{\text{SG}_i} - \theta, 0)
\end{equation}
Intuitively, a high drop ratio indicates that the \codename score decreases immediately after the peak at position~$i$, without progressively unraveling the surprising plot.

\begin{figure*}[!ht]
\centering
    \begin{subfigure}[b]{0.45\textwidth}
        \includegraphics[width=\textwidth]{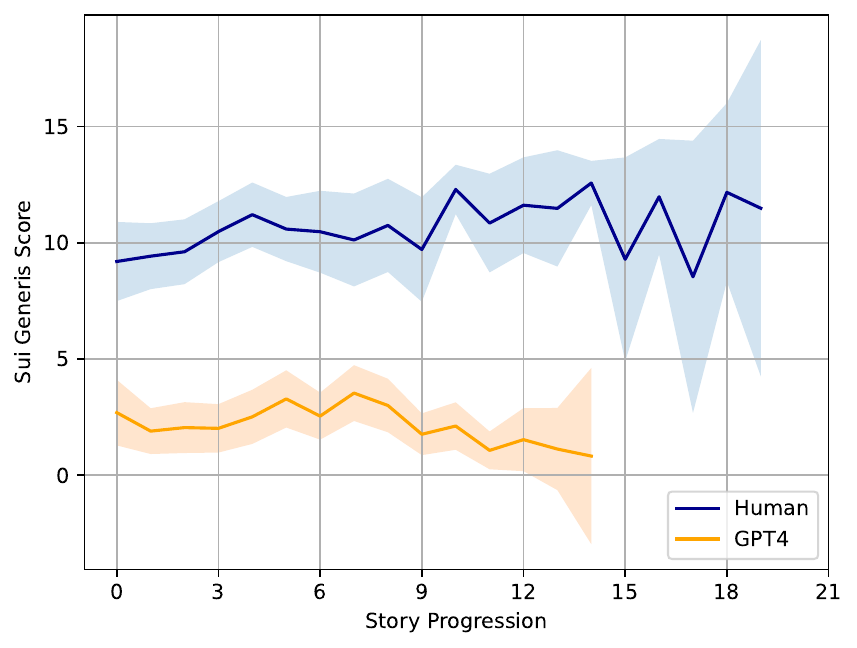}
        \caption{WritingPrompts -- Human vs. GPT-4}
    \end{subfigure}
    \begin{subfigure}[b]{0.46\textwidth}
        \includegraphics[width=\textwidth]{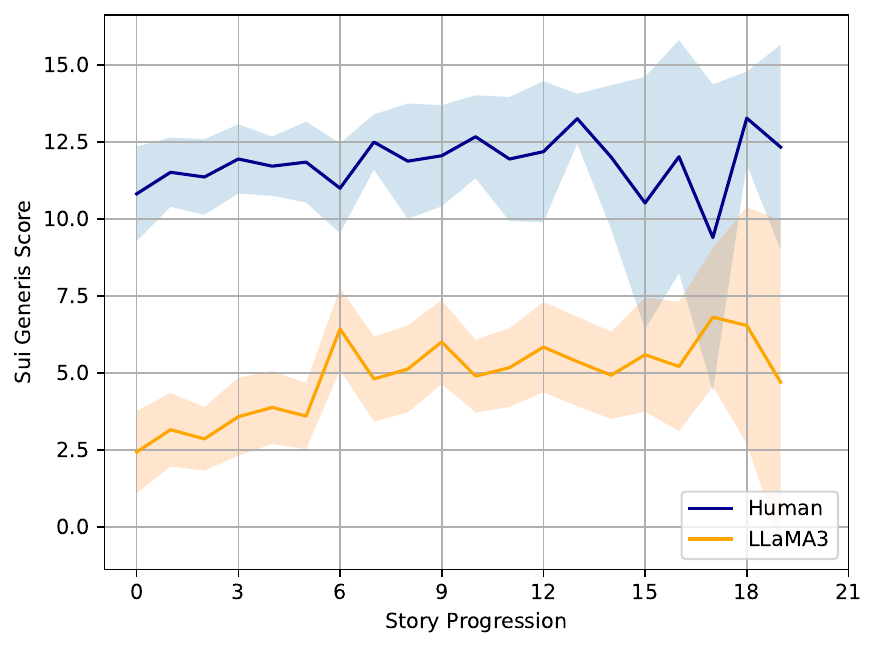}
        \caption{WritingPrompts -- Human vs. LLaMA-3}
    \end{subfigure}\\
    \begin{subfigure}[b]{0.45\textwidth}
        \includegraphics[width=\textwidth]{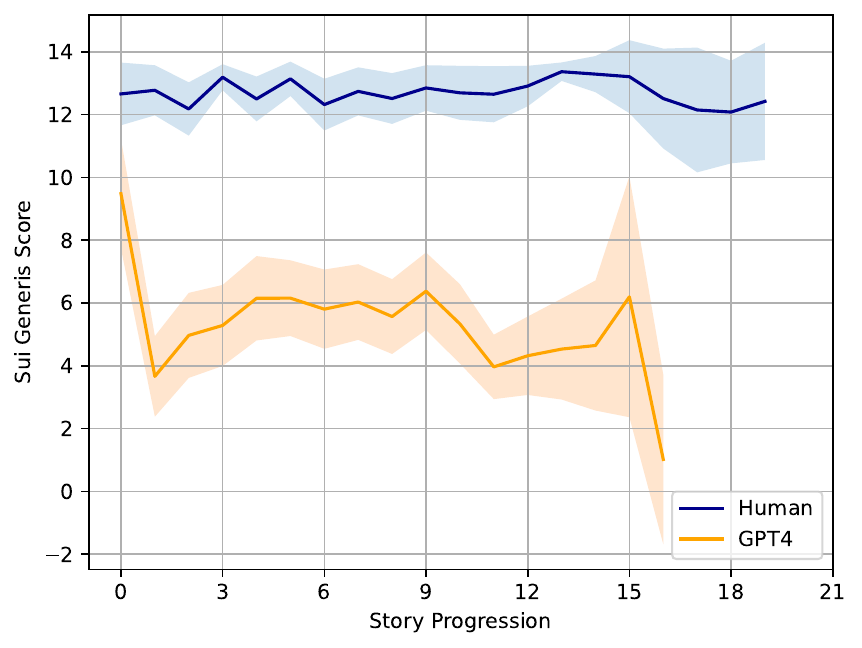}
        \caption{Wiki -- Human vs. GPT-4}
    \end{subfigure}
    \begin{subfigure}[b]{0.45\textwidth}
        \includegraphics[width=\textwidth]{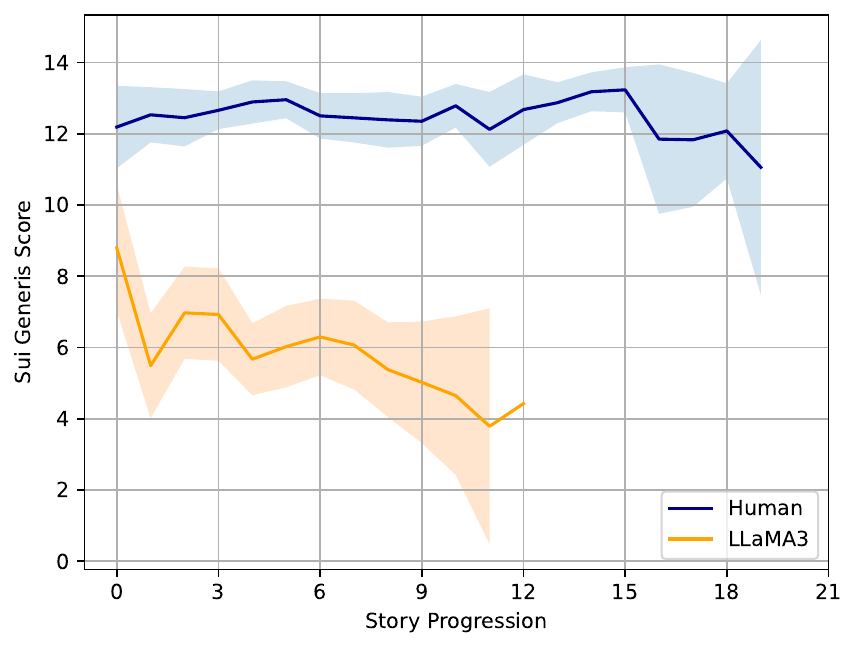}
        \caption{Wiki -- Human vs. LLaMA-3}
    \end{subfigure}
\caption{Average \codename scores of story segments generated by humans versus GPT-4~(in (a) and (c)) or LLaMA-3~(in (b) and (d)) at varying segment positions~(\textit{x-axis}). The shaded area represents the confidence interval with confidence level~$\gamma=0.95$.}
\label{fig:average_line}
\end{figure*}

\section{Experimental Setup}

\subsection{Datasets}
We test our \codename score on~100 stories from two story datasets:~1) the WritingPrompts dataset~\citep{fan-etal-2018-hierarchical}, which contains story prompts and the corresponding human-written stories posted on an online forum\footnote{\url{https://www.reddit.com/r/WritingPrompts/}} by the year of~2018, and 2) plot summaries of TV episodes in the year of~2023 crawled from Wikipedia.

For WritingPrompts, we randomly sample~$50$ stories and directly use the story prompts by humans for LLM story generation.\footnote{The prompt templates used for LLM generation are listed in Appendix. We also tried using GPT-4 to generate 20 different instruction prompts and evaluated the stories generated using these prompts. Results show very small variance~(with a standard deviation of~1.1) in \codename scores among stories generated with different instruction prompts.} To account for the fact that the paragraph segmentation of stories generated by different human authors and LLMs can vary significantly, thus affecting the segment-level scores, we break both human- and LLM-generated stories into segments of~50 words.\footnote{We chose word-level segmentation so that information is distributed evenly across segments. And the average number of words in each paragraph of WritingPrompts stories is around~50.} Examples of such segments are shown in Figure~\ref{fig:echo_example} within square brackets [...]. This results in a total of~$1,887$ segments in the original and LLM-generated stories for WritingPrompts.

On~50 randomly sampled Wiki plot summaries, we use the first two sentences from each plot summary as the prompt for LLM story generation. We segment both human and LLM stories by a fixed length of~30 words, which is the average length of a sentence in Wiki plot summaries.\footnote{Because the Wiki plot summaries are more condensed than WritingPrompts stories, and usually each sentence can cover very rich plot.} We test on a total of~$1,862$ segments from the original and LLM-generated stories.

\subsection{\codename Scoring Setup}
\looseness=-1
We sample~$K=20$ alternative continuations at each position in the story. We set~$\lambda=0.9$ when computing the weighted average of the negative log of echoes for the \codename score in Eq.\ref{eq:echo_score}. To compute the drop ratio, we set the threshold~$\theta=0.5$.

\subsection{Large Language Model Setup}
\label{sec:llm_setup}
We use GPT-4~\citep{openai2024gpt4technicalreport} and LLaMA-3-70B-Instruct~\citep{dubey2024llama3herdmodels}~(more details in the Appendix) with~$top\_p = 1.0$, sampling temperature~$\tau = 1.0$ for story generation.\footnote{As \cite{peeperkorn2024temperaturecreativityparameterlarge} find that the influence of temperature on the creativity of LLMs is weak.}
When computing the \codename score, we use the same story generation model~(with~$top\_p = 0.5$ and~$\tau = 1.0$) to generate alternative continuations~$C_j$, and GPT-4 with~$top\_p = 0.5$ and~$\tau = 0$ to estimate the function~$a(s, C)$~(plot entailment), i.e. if similar plot in the original story segment appears in the alternative continuations.

\section{Results}

\begin{figure}[!ht]
    \centering
    \begin{subfigure}[b]{0.38\textwidth}
        \includegraphics[width=\textwidth]{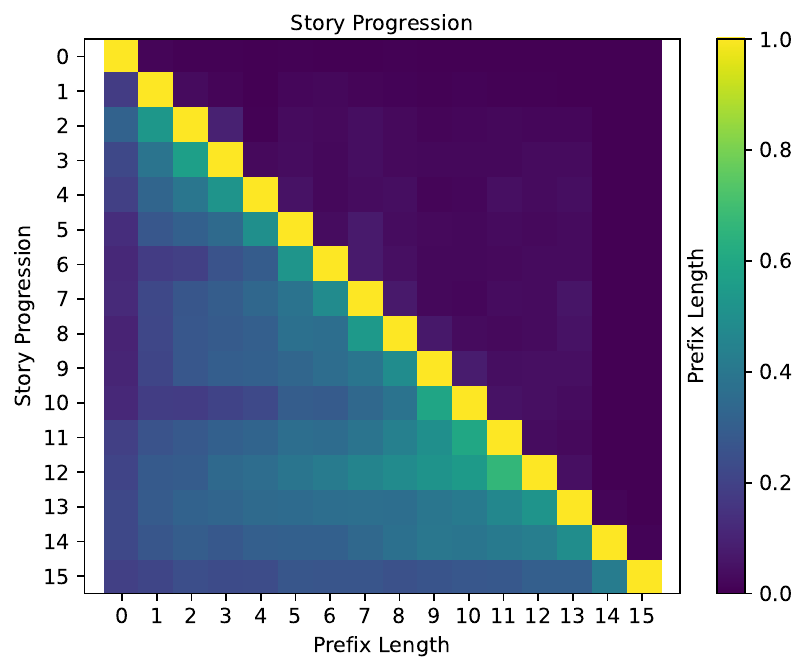}
        \caption{Wiki -- Human vs. GPT-4}
    \end{subfigure}
    \begin{subfigure}[b]{0.38\textwidth}
        \includegraphics[width=\textwidth]{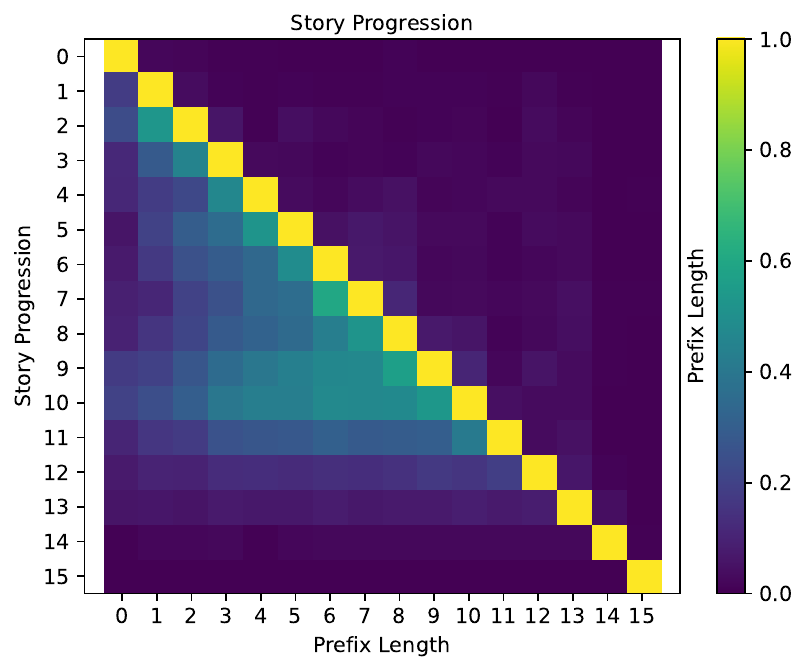}
        \caption{Wiki -- Human vs. LLaMA-3}
    \end{subfigure}
\caption{Heatmap of the average Echo scores~$p_{i, j}$ for the~$i$-th story segment given story prefix~$S_{-j}$ on~50 Wiki stories. The lower left triangle shows~$p_{i, j}$ on LLM~(GPT-4 or LLaMA-3) generated stories, while the upper right triangle shows the transpose matrix of~$p_{i, j}$ on human stories.}
\label{fig:pij_heatmap_main}
\vspace{-10pt}
\end{figure}

\begin{figure}[!ht]
    \centering
    \includegraphics[width=0.38\textwidth]{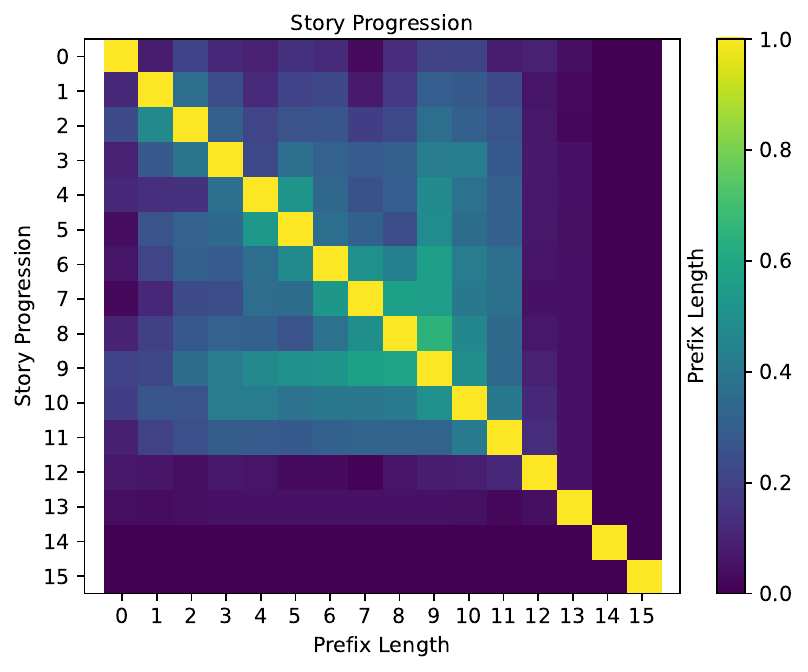}
\caption{Heatmap of the average Echo scores~$p_{i, j}$ for the~$i$-th story segment given story prefix~$S_{-j}$ on~20 Wiki stories by LLaMA-3. The lower left triangle shows the original Echo scores~$p_{i, j}$, while the upper right triangle shows the transpose matrix of cross-model Echo scores~$p_{i, j}$ (using GPT-4 to generate alternative continuations).}
\label{fig:pij_heatmap_crossmodel}
\vspace{-10pt}
\end{figure}

\paragraph{\codename Score: Human vs. LLMs}
We first compare the average \codename scores of story segments produced by humans versus LLMs~(including GPT-4 and LLaMA-3). As shown in Figure~\ref{fig:average_line}, for both datasets, human-written stories yield significantly higher \codename scores than LLM-generated ones. We also show the average Echo score matrix~$p_{i, j}$ on Wiki data in Figure~\ref{fig:pij_heatmap_main}~(see the other heatmap figures in the Appendix). For LLM-generated stories, the Echo rate increases as more of the previous plot is revealed~(i.e., as the prefix length increases), while for human-written stories, the LLM can only occasionally predict the beginning section of the story. These results indicate that LLM-generated stories are largely composed of idiosyncratic plot elements that are echoed across multiple LLM generations, while human-written stories often contain plot elements that land far in the tail -- or not at all -- of LLMs' output distribution.\footnote{We further discuss the memorization effect in Section~\ref{sec:limitations}.} 

\paragraph{\codename Score: Cross-Model Scoring} 
Next, we test if there are common echoes across different LLMs. To this end, we compute the cross-model \codename scores \---\ scores of LLaMA-3 generated stories given the alternative continuations generated by GPT-4. Results on~20 Wiki and~20 WritingPrompts stories show that the cross-model \codename scores are slightly higher than the original scores, but the differences are small~(0.6 on Wiki stories and~1.6 on WritingPrompts) compared to the gap between human and LLaMA-3 story scores~($>6.0$). As shown in Figure~\ref{fig:pij_heatmap_crossmodel}, the cross-model Echo scores distribute similarly to the original Echo scores. These results indicate that the plot elements generated by LLaMA-3 are echoed not only in its own generations, but also in the alternative continuations generated by GPT-4. In other words, the plots generated by LLMs are echoed frequently across different models, while human-generated plots are rarely echoed by any of these LLMs.

\paragraph{Drop Ratio: Human vs. LLMs} 
As shown in Table~\ref{tab:drop_ratio}, LLM-generated stories lead to~7--9 higher drop ratio than human-written ones. This indicates that, in LLM-generated stories, there are more cases where the \codename score drops immediately after a peak, while in human-written stories, the score often peaks and then stays high. We further show that this is related to the overly fast pacing in LLM-generated stories in Section~\ref{sec:discussion}.

\begin{figure}[!ht]
    \centering
    \includegraphics[width=.42\textwidth]{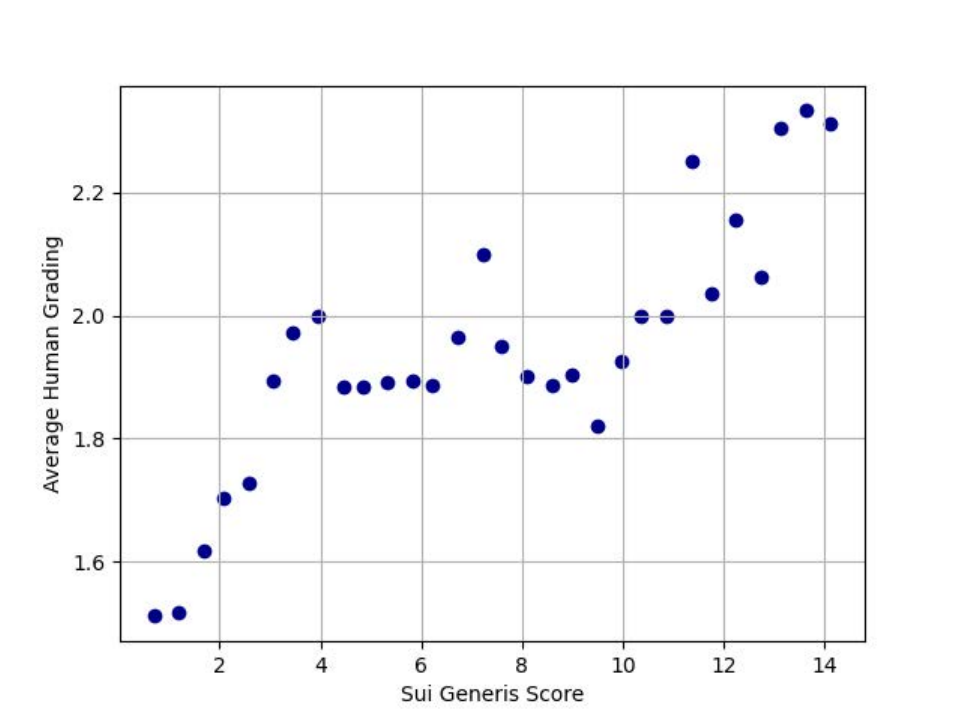}
\caption{Average human grading of surprisal versus the \codename scores on story segments generated by LLaMA-3 based on the prompts from WritingPrompts. The data points are binned by an interval of around~1.4 based on the \codename scores.}
\label{fig:human_eval}
\vspace{-10pt}
\end{figure}

\paragraph{\codename scores correlate with human judgment of surprise level}
We invite four human judges, fluent in English, to evaluate the surprise level of story segments in~9 randomly selected stories~(159 segments) generated by LLaMA-3 based on the story prompts from WritingPrompts. The study was approved by Microsoft Research's Institutional Review Board~(case number~11035). Each human annotator in the study had signed a consent form electronically, informing them of the task and potential risks. Specifically, we ask each human annotator to read through each story segment by segment and annotate how surprising each segment is from level~1--3~(1-most-anticipated, 2-neutral, or 3-most-surprising) based on the story so far. To enhance agreement among annotators, we give them~5 examples~(excluded from the formal annotation task) and ask them to discuss their annotations to reach an agreement prior the formal annotation task. For the formal annotation task, the inter-annotator agreement based on Krippendorff's alpha is moderate~(0.68).
Results show that segment-level \codename scores correlate moderately with the average human judgment of surprise level: the Spearman's rank correlation between \codename scores and the average human rating is significant\footnote{We use the p-value threshold~$p=0.05$ for all the significance test in this paper unless noted otherwise.} and the magnitude is~0.55. The correlation between \codename scores and the average human rating is even stronger than that between each individual human rating and the average among the other three annotators~(with Spearman's correlation between~0.38 and~0.49). The relationship between human ratings and \codename scores is visualized in Figure~\ref{fig:human_eval}, where average human ratings (y axis) are shown for segments whose \codename scores fall within a bin within a 1.4 interval around the x axis value. While humans rate segments with a wide range of mid-level \codename scores as mildly surprising, they agree with \codename on segments with highest and lowest scores. 

\begin{table}[!t]
    \centering
    \begin{tabular}{l|rr}
    & Human & LLM \\\midrule
    WP: Human vs GPT-4 & 3.7 & 11.3 \\
    WP: Human vs LLaMA-3 & 1.7 & 9.0 \\
    Wiki: Human vs GPT-4 & 0.4 & 9.1 \\
    Wiki: Human vs LLaMA-3 & 0.5 & 7.8 \\\midrule
    \end{tabular}
    \caption{Comparing the drop ratio (in percentage) of stories generated by humans versus GPT-4 or LLaMA-3 on WritingPrompts~(\textit{WP}) and Wiki data. LLM-generated stories obtain consistently higher drop ratio than human-written ones.}
    \label{tab:drop_ratio}
    \vspace{-10pt}
\end{table}

\paragraph{The story prompt itself affects \codename scores for both human and LLM stories} 
\looseness=-1
We hypothesize that the story prompt will have similar impact on the \codename scores of stories generated by humans and LLMs, as certain story ideas in the prompt (e.g. in Figure~\ref{fig:echo_example}(c)) simply lead themselves to more interesting and diverse continuations. To test the hypothesis, we compute the Spearman's correlation between the \codename scores of pairs of human-written and LLM-generated stories under the same prompt. The correlation between the scores of human- and GPT-4-generated stories is~0.48 with p-value~$<0.01$, which indicates a moderate correlation. The correlation between human- and LLaMA-3-generated stories is~0.34 with p-value~$<0.01$, which indicates a weak but significant correlation. This suggests that, although human-written stories are scored significantly higher than LLM-generated ones, story prompts that lead to relatively low-scored stories by LLMs also lead to low-scored stories by humans. 
Thus, one can boost the \codename score of human and LLM-generated stories by improving the story prompts; prompting is itself a creative endeavor.

\paragraph{\codename captures plot-level surprisal beyond token likelihoods}
We further investigate how robust the \codename score is to paraphrasing that would affect token-level surprisal measurements such as perplexity, even when there is no significant difference at the more abstract narrative level. To this end, we randomly sample~10 human-written stories and prompt LLaMA-3 to paraphrase each story without changing the semantics~(which we manually verified). %
Next, we compute the \codename and perplexity scores for each of the original and paraphrased stories. Results show that the \codename score remains largely unchanged between the original and paraphrased stories~(with a~4\% drop on average after being paraphrased), while the perplexity score drops by~15\% after being paraphrased. Additionally, the ranking of the perplexity scores among the~10 stories changes greatly after the stories are paraphrased: the Spearman's rank correlation between the perplexity scores on the original and paraphrased stories is~0.55, and the correlation is non-significant. By contrast, the Spearman's rank correlation between the \codename scores on the original versus paraphrased stories is both significant and much stronger~(0.75). These results show that the \codename score is far less sensitive to local wording than token-level perplexity, instead capturing plot-level surprise.

\section{Discussion}
\label{sec:discussion}

\paragraph{Comparison with Other Similarity Metrics}
We compare our prompting-based plot entailment assessment with other similarity/diversity metrics by measuring how these metrics correlate with human judgments. We conduct a human evaluation in which we invite four human judges~(who are fluent in English) to judge 20 randomly sampled pairs of plot segments. The study was approved by Microsoft Research's Institutional Review Board~(case number~11035). Each human annotator in the study had signed a consent form electronically, informing them of the task and potential risks. We give each human judge the same prompt as GPT-4 (see Fig. A.1 and Fig. A.2 and ask them to provide their answers among \textit{yes}, \textit{no}, or \textit{partially}.\footnote{The inter-annotator agreement based on Fleiss’s Kappa is fair~(0.33).} We then convert their answers to prediction scores between~0 and~1~(yes$\rightarrow$1, no$\rightarrow$0, partially$\rightarrow$0.5) and take the average score among the four judges on each example. Next, we compute the Spearman's correlation between the average human score and our prompting-based assessment, comparing it against five other text similarity/diversity metrics as recommended in \cite{shaib2024standardizingmeasurementtextdiversity}: compression ratio, self-BLEU~\cite{Zhu2018TexygenAB}, n-gram diversity~\cite{padmakumar2024does}, homogenization score~(based on ROUGE-L)~\cite{padmakumar2024does} and embedding-based similarity~\cite{padmakumar2024does,wang2024guidingdiversifyingllmbasedstory}.\footnote{We obtain sentence embeddings from the \textit{all-MiniLM-L6-v2} model.}
Results show that our prompting-based entailment assessment score correlates very strongly with the average human score with a Spearman's correlation of~0.85. The compression ratio, self-BLEU and n-gram diversity correlate weakly with human score with Spearman's correlation of~0.07, 0.33 and 0.23, respectively. These metrics rely mostly on surface-form features like ngram overlaps but fail to capture sentence-level semantics. The homogenization score and embedding-based similarity correlate moderately with the average human score~(Spearman's correlation:~0.46 and~0.50) but not as strongly as our prompting method. These results demonstrate that our prompting method is better at measuring plot entailment than existing similarity metrics. In addition, the prompting method gives us more control over the level of similarity that we want to capture, while it is difficult to control or understand the level of similarity captured by the embedding-based score.

\paragraph{What are the main characteristics of low-scored stories?} 
We examine the lowest-scored stories generated by humans and LLMs. These stories typically fall into two categories. The first includes stories that are too bland given the story prompts (see Table~D.1 in the Appendix), which is more common in LLM-generated stories than human-written ones.
The second class of low-scoring stories do exhibit rich plots, but the plot elements themselves are banal and are used repetitively across many LLM generations~(see the examples in Table~D.2 in the Appendix).

\paragraph{What are the main characteristics of high-scored stories?} 
We further examine the highest-scored stories from both humans and LLMs. The common features across human- and LLM-generated stories are that they usually contain non-traditional plots with rich details~(see the example in Table~D.4 in the Appendix).
Additionally, human-written stories offer more interesting twists and non-linear storytelling~(e.g. clues, distractions, interlude and flashbacks) than LLM-generated ones, as shown by Table~D.3 in the Appendix.

\paragraph{Segment-level \codename scores can be used to identify key plot elements} 
We further examine segment-level \codename scores and find that the high scores usually correspond to the key narrative developments and turning points. For example, in Table~\ref{tab:mixed_score_human_examples}, the high-scored segments correspond to the key plot turns~(e.g. when characters find that everyone has received the same note under their doors, and the twist ending where the last person who opened her door revealed the note she received), while the low-scored segment is a predictable consequence of the previous developments.

\paragraph{High drop ratio in LLM-generated stories} 
A user study involving expert writers ~\citep{Chakrabarty2024} suggests that
 LLM-generated stories differ from human-written ones in narrative pacing.
We therefore examine whether the high drop ratio in LLM-generated stories compared to human-written ones is related to this phenomenon. Indeed, we find that the high drop ratio in LLM stories is typically 
associated with overly-hasty narrative pacing in LLM-generated stories, often leading to abrupt resolutions that leave key suspenseful plot elements unresolved. 
The result is a drop in the \codename score ~(see the example in Table~D.5 in the Appendix).
By contrast, human-written stories tend to have slower pacing, introduce suspenseful elements more gradually, and provide a final resolution that ties up hanging plot threads~(see the human-written story under the same prompt in Table~D.3 in the Appendix).

\paragraph{Computational Cost}
The number of LLM calls needed to compute the \codename score for a story~$S$ is~$[n + K \cdot n (n - 1) / 2]$, where~$K$ is the number of alternative continuations~$C_j$ sampled given each prefix~$S_{-j}$, and~$n$ is the total number of segments in the story~$S$. In our experiment, running \codename for an average-length story~(10 segments, 500 words) takes~$910$ LLM calls, which amounts to around~$700k$ input tokens and~$700k$ output tokens through an LLM. This will cost around~\$7 USD using GPT-4.1~(one of the best LLMs) based on current pricing.\footnote{\url{https://openai.com/api/pricing/}} We anticipate that this cost will decrease rapidly in the near term given current trends.\footnote{\url{https://epoch.ai/data-insights/llm-inference-price-trends}}

\paragraph{Possible Uses of \codename In Generation}
Beyond evaluation, the \codename score can also be used to improve diversity through training or inference-time optimization. For instance, we can generate a story fragment by fragment, where at each step, we sample~$M$ alternative segments and keep the one with the highest \codename score. We test this approach by sampling a story~(\emph{S-100}) using GPT-4 with~$M=100$ given a prompt from WritingPrompts. \emph{S-100} obtains~+5.3 higher \codename score and is more interesting than the story generated given the same prompt in one pass~(see the stories in the Appendix), at the cost of 2000 times more LLM calls. \emph{S-100} even achieves human-level \codename score when scoring with \emph{SG-20}~(where we set the number of alternative continuations~$K$ to~20). However, by scaling up~$K$ from~20 to~100~(which increases the resolution at which we can capture unique segments), we can still detect the gap between human and LLM-generated stories \---\ \emph{S-100} obtains an \emph{SG-100} score of~3.8, which is~6.4 points lower than the \emph{SG-100} score of the human-written story given the same prompt. Note that computing the \emph{SG-100} score is 20 times less expensive than generating \emph{S-100}. In general, the number of LLM calls required for computing \emph{SG-K} is linear in~$K$, while the number of LLM calls for generating a story \emph{S-M} that maximizes \emph{SG-K} is proportional to~$K \cdot M$.

\section{Limitations}
\label{sec:limitations}
The WritingPrompts corpus and part of the Wiki data used in the experiment may have been in the training corpora of the LLMs evaluated. Although our results show no effect of a memorization effect for the stories we tested, it is possible that this kind of training effect might surface for other datasets that were included in the LLMs' training corpora. As the ability of LLMs to memorize long texts improves, the \codename score will be applicable only to corpora that were not included in the training data. 
Additionally, the scoring method relies on the capability of an LLM to identify semantic similarities at the desired level. In our experiment, we always use GPT-4 for the entailment judgment, as our human evaluation shows that it is capable of mimicking human judgment of entailment accurately. If one were to use our metric with a different entailment judgment model, they would first need to ensure that this model is at least as capable as GPT-4. Substituting a less capable model and using it for both generation and entailment judgment could lead to a bias, which could produce unreliable scores.

\section{Conclusion}
\looseness=-1
We introduced the \codename score to quantify the uniqueness of LLM-generated stories at a narrative level when compared to their alternative generations. Our experiments on 100 LLM- versus human-written stories demonstrated the lack of plot-level diversity in LLM-generated stories. Regardless of the LLM used to generate them, these stories are often composed of echoic plot elements, in contrast to the more varied and unique elements found in human-written stories. Moreover, our human evaluation showed a moderate correlation between \codename scores and human judgments of surprise, suggesting that the \codename score can serve as a useful tool for assessing the uniqueness and surprise level of LLM-generated content. Furthermore, the \codename score has the potential to be extended to assess the uniqueness of sequential content across different languages and for other modalities beyond text. For example, it could be adapted to assess the uniqueness of compositions in music, visual elements in images, and temporal segments in video content by substituting the entailment function with similarity measurement in the new modality.

This work also sheds light on the societal impact of AI-generated content, as it suggests that disseminating AI-generated content at scale while being unaware of LLMs' tendency to homogenize might negatively affect cultural diversity, reducing the collective creativity of online content and the richness of educational experiences. This is reminiscent of how the lack of diversity in recommender systems has led to social polarization and has amplified social bias~\cite{BOJIC2024103383}.
On the bright side, the \codename score holds promise for future work aimed at enhancing diversity in AI- or human-AI-generated content. For instance, it can be integrated with inference optimization algorithms to produce more unique outputs. Additionally, the score can help identify low-scoring segments in AI-generated content, where human input can be introduced to enrich creativity and originality, as \cite{Anderson2024Homogenization} suggests that by informing users that a particular piece of model output is homogeneous, it may help users resist model-induced homogenization at the group level.

\paragraph{Data, Materials, and Software Availability.} The code for computing the Sui Generis score is deposited in Github~\cite{SuiGenerisGithub}.

\vspace{20pt}
\bibsplit[14]

\bibliography{custom}

\end{document}